\documentclass[letterpaper]{article} 
\usepackage{aaai24}  
\usepackage{times}  
\usepackage{helvet}  
\usepackage{courier}  
\usepackage[hyphens]{url}  
\usepackage{graphicx} 
\usepackage{natbib}  
\usepackage{caption} 
\usepackage{algorithm}
\usepackage{algorithmic}

\usepackage{booktabs}
\usepackage{amssymb}
\usepackage{pdfpages}

\usepackage{newfloat}
\usepackage{listings}
\DeclareCaptionStyle{ruled}{labelfont=normalfont,labelsep=colon,strut=off} 
\floatstyle{ruled}
\newfloat{listing}{tb}{lst}{}
\floatname{listing}{Listing}
\title{NeuSurf: On-Surface Priors for Neural Surface Reconstruction from Sparse Input Views}
\author {
    Han Huang\textsuperscript{\rm 1,\rm 2},
    Yulun Wu\textsuperscript{\rm 1,\rm 2},
    Junsheng Zhou\textsuperscript{\rm 1,\rm 2},
    Ge Gao\textsuperscript{\rm 1,\rm 2}\thanks{Corresponding author.},
    Ming Gu\textsuperscript{\rm 1,\rm 2},
    Yu-Shen Liu\textsuperscript{\rm 2}
}
\affiliations {
    \textsuperscript{\rm 1}Beijing National Research Center for Information Science and Technology (BNRist), Tsinghua University, Beijing, China\\
    \textsuperscript{\rm 2}School of Software, Tsinghua University, Beijing, China\\
    \{h-huang20, wu-yl22, zhoujs21\}@mails.tsinghua.edu.cn, \{gaoge, guming, liuyushen\}@tsinghua.edu.cn
}

\usepackage{bibentry}

\begin{document}

\maketitle

\begin{abstract}
Recently, neural implicit functions have demonstrated remarkable results in the field of multi-view reconstruction. However, most existing methods are tailored for dense views and exhibit unsatisfactory performance when dealing with sparse views. Several latest methods have been proposed for generalizing implicit reconstruction to address the sparse view reconstruction task, but they still suffer from high training costs and are merely valid under carefully selected perspectives. In this paper, we propose a novel sparse view reconstruction framework that leverages on-surface priors to achieve highly faithful surface reconstruction.  Specifically, we design several constraints on global geometry alignment and local geometry refinement for jointly optimizing coarse shapes and fine details. To achieve this, we train a neural network to learn a global implicit field from the on-surface points obtained from SfM and then leverage it as a coarse geometric constraint. To exploit local geometric consistency, we project on-surface points onto seen and unseen views, treating the consistent loss of projected features as a fine geometric constraint. The experimental results with DTU and BlendedMVS datasets in two prevalent sparse settings demonstrate significant improvements over the state-of-the-art methods. 
\end{abstract}

\section{Introduction}
Surface reconstruction from multi-view images is a critical task in the fields of computer vision and computer graphics. Traditional methods, like Multi-View Stereo \cite{hypotheses,colmap,mvsnet}, leverage geometric consistency between images to compute the depth map. Subsequently, they obtain the reconstructed point cloud through depth map fusion. Nonetheless, the conversion of this intermediate representation might introduce cumulative geometric errors. In scenarios with sparse views, the MVS method faces challenges in reconstructing a smooth and detailed surface due to the scarcity of matching points and variations in viewing angles.

\begin{figure}[!t]
\setlength{\belowcaptionskip}{-0.375cm}

    \centering
    \includegraphics[width=0.47\textwidth]{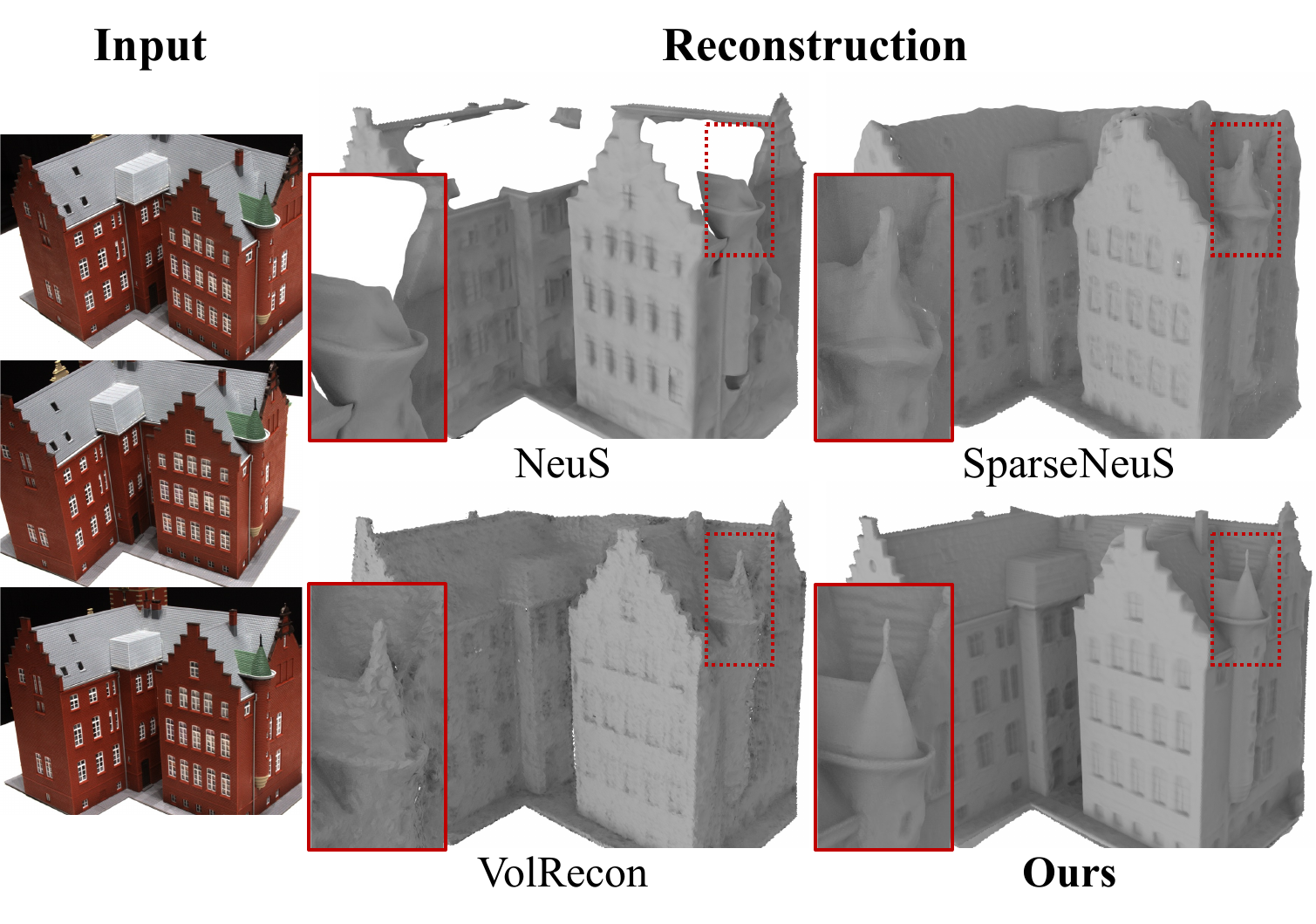}
    
    \caption{ Surface reconstruction results from sparse-view images (large-overlap setting).  The state-of-the-art methods SparseNeuS~\cite{sparseneus} and VolRecon \cite{volrecon} produce noisy and broken surfaces, while the results of ours (NeuSurf) are detailed and complete. } 

\label{fig:teaser}
\end{figure}
In recent years, neural rendering-based surface reconstruction methods \cite{volsdf,neus,monosdf} have been widely used to improve the reconstruction results by producing smoother and more complete geometries. These methods simultaneously optimize both implicit geometry and neural radiance fields by minimizing the discrepancy between the rendered and the ground truth images. However, the well-optimized photometric loss may distort the geometry due to shape-radiance ambiguity \cite{vdnnerf}. Especially under the situation of sparse view input, learning of the geometry field may further collapse.

As a remedy, some recent methods \cite{sparseneus,volrecon} partly solved the sparse view reconstruction by introducing additional generalizable priors. They first learn geometric priors from large-scale data, and then fine-tune the implicit fields to achieve surface reconstruction in new scenarios. However, the learned priors are only effective in a fixed large-overlap sparse setting. Once the sparse view is inconsistent with the pose distribution with the fixed setting, the priors will be invalid and fail to bring robust guidance to surface reconstruction. As a result, the performance of the generalizable methods is dramatically limited in cases with different sparse settings. Due to the long training time and complex data pre-processing, it is unrealistic to train a prior for each sparse setting.

In this paper, we propose a novel sparse view reconstruction framework to achieve highly faithful surface reconstructions using on-surface point priors. The proposed priors are achieved directly from the raw input sparse views without requiring any extra training or data, which effectively improve the reconstruction results and are robust to different sparse settings. The effectiveness of our method is not affected by different sparse settings. Specifically, we obtain initial on-surface points by SfM method, which can be considered as ``free data'' in the training process. Instead of using these points directly for depth supervision, we design two constraints in terms of loss functions to make full use of on-surface points. One is conducted with the guidance of the global geometry field, where we train a neural network to learn the geometric field of on-surface points and then use that field as a rough geometric constraint. The other is local geometric refinement loss, which is achieved by projecting surface points onto visible and invisible views and optimizing the consistency of projection features to reconstruct fine local geometry. Our contributions are listed below.
\begin{itemize}
    \item We propose a novel framework for surface reconstruction from sparse view images. our framework takes full advantage of on-surface point clouds, which is easy to access, as an additional effective supervision to guide the geometry learning.
    \item We use the global geometric fields obtained from the surface points to help learn rough and continuous geometry. In addition, we optimize the local feature consistency of on-surface points to help learn fine geometry.
    \item We achieve state-of-the-art reconstruction results under different prevalent sparse settings on the widely-used DTU and BlendedMVS datasets.
\end{itemize}

\begin{figure*}[!ht]
    \centering
    \includegraphics[width=\textwidth]{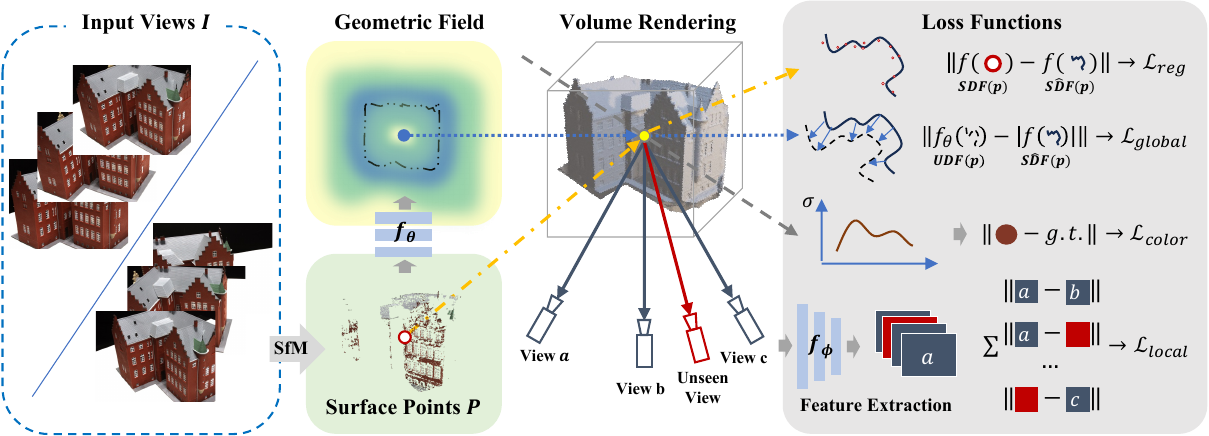}
    \caption{
    Structure of NeuSurf. For a set of $3$ source views (in large-overlap or little-overlap), we first obtain the surface points by the SfM method. Within the on-surface points, we train a UDF network as the geometric field and leverage it as global geometry alignment. Then we utilize the feature consistency between seen and unseen views to optimize the local geometry. In addition to the RGB rendering loss, explicit on-surface points regularization can be improved as an additional loss.
    }
\label{fig:pipeline}
\end{figure*}

\section{Related Work}
\subsection{Multi-View Stereo (MVS)}
Traditionally, MVS methods use point clouds \cite{pointbased1,pointbased2}, depth maps \cite{depth1,depth2,depth3} and voxel grids \cite{voxel1,voxel2,surfacenet} as 3D representations of scenes to reconstruct the surface geometry. Due to the need for parallel computing, depth maps based methods are now widely used. Depth maps based methods predict the depth of each image and then fuse them to get the surface point cloud of the object. After the point cloud is obtained, Screened Poisson surface reconstruction method \cite{possion} can be used to further obtain the surface mesh.

\subsection{Neural Implicit Representations}
Recently, advanced methods employing neural implicit functions to represent 3D scenes have emerged, and these can be applied to shape representation \cite{zhou20223d,occupancy,zhou2023levelset}, novel view synthesis \cite{nerf,neuralvoxel,wenyuan} and multi-view 3D reconstruction \cite{unisurf,neus,volsdf,monosdf,ma2023towards}. Given raw point clouds, Neural-pull \cite{neuralpull} and CAP-UDF \cite{capudf} design neural networks to learn geometric fields that represent 3D shapes. They provide a way to transform the raw point cloud representation of an object surface into a geometric field representation without the ground truth values of the geometric field. Neural Radiance Fields (NeRF) \cite{nerf}, as a popular novel view synthesis method in recent years, encodes color fields and volume density fields with implicit representations. 
\subsection{Neural Surface Reconstruction}
Inspired by NeRF, NeuS \cite{neus} and VolSDF \cite{volsdf} were
proposed to encode signed distance function (SDF) and color fields of the scene. By minimizing the discrepancy between the rendered image and the ground truth image, they can obtain a smooth and complete SDF geometric field. To make the geometry of the SDF field more precise, MonoSDF \cite{monosdf} and Geo-NeuS \cite{geoneus} add geometric loss in addition to photometric loss, which reduces the possible bias in the volume rendering process. The above-mentioned methods are all based on dense view input. However, in real world, there are often only fewer views that can be used for reconstruction.

To achieve sparse view reconstruction, SparseNeuS \cite{sparseneus} and VolRecon \cite{volrecon} learn generalizable geometric priors from large-scale data, and then fine-tune on new scenes. They train on 75 scenes of the DTU \cite{dtu} for several days and then test on the remaining 15 scenes. Although they have obtained some geometric priors of the data sets through large-scale training, they still only generalize in the case of a fixed sparse setting.

In observation, when only sparse views are given, the complexity of the neural surface learning increases, and it is more likely to achieve a collapsed geometry (incomplete, noisy), as shown in Figure \ref{fig:teaser}. Current neural surface learning methods with large-scale training priors are often time-intensive and only useful within a specific sparse setting.

Unlike previous works, instead of using costly large-training priors, our method attempts to exploit the priors of surface points to optimize the neural surface representation both globally and locally.

\section{Method}

Given sparse-view images $I = \{I_i|i\in1,..,\mathcal{M}\}$ with camera poses $T = \{T_i,|i\in1,...,\mathcal{M}\}$, our goal is to reconstruct the high-quality geometry $S$ of the scene represented by $I$. In this paper, we propose NeuSurf, a neural surface reconstruction method with sparse input views, as illustrated in Figure \ref{fig:pipeline}.

Our motivation for proposing NeuSurf is to reduce the complexity of neural surface learning with non-training priors and produce more complete and detailed reconstruction. Specifically, points obtained by Structure from Motion \cite{colmap} are regarded as a ``free'' data source as it is easy to acquire with no extra input. We denote the surface points obtained by SfM as $P$, which we do not use as a depth loss function for the neural rendering directly. Instead, we learn a global geometric field $f_\theta$ from these on-surface points and use it to align the rough geometry. To get fine details of the surface, local feature consistency for on-surface points is optimized as another constraint.  

\subsection{Learning Neural Implicit Surface by Volume Rendering}
We represent the geometry and appearance with SDF fields and color fields, which are learned by the neural rendering pipeline. We adopt NeuS \cite{neus} as our baseline, which defines the geometry as the zero-level set of signed distance function (SDF) $S = \{\mathbf{x}\in {\mathbb{R}}^3| f(\mathbf{x}) = 0 \}$, and develop a novel volume rendering method to learn the geometry and appearance of the scene. The SDF and color are parameterized with two MLPs as provided by NeuS.

Given a pixel from one image, the ray could be denoted as $\{\mathbf{r}(t) = \mathbf{o}+t\mathbf{d}|t>0\}$, where $\mathbf{o}$ is the camera center and $\mathbf{d}$ is the direction of the ray. The rendered color is accumulated by volume rendering with $N$ discrete points:
\begin{equation}
    C(\mathbf{r}) = \sum_{i=1}^NT_i\alpha_ic_i,
\end{equation}
where $T_i$ is the accumulated transmittance, $\alpha_i$ is opacity values, as denoted by
\begin{equation}
    T_i = \prod_{j=1}^{i-1}(1-\alpha_i),
\end{equation}
\begin{equation}
    \alpha_i = \max \left (\frac{\Phi_s(f(\mathbf{r}(t_{i}))) - \Phi_s(f(\mathbf{r}(t_{i+1})))}{\Phi_s(f(\mathbf{r}(t_i)))}, 0 \right).
\end{equation}
$\Phi_s$ follows NeuS, expressed as $\Phi_s(x) = (1+e^{-sx})^{-1}$ with $s$ being a trainable, diminishing parameter.
\subsection{On-Surface Global Geometry Alignment}
On-surface points are discrete and thus fail to determine some surface locations. We attempt to learn a continuous geometric field from the discrete point cloud as a prior to provide coarse information for surface learning. With the guidance of the prior field, we largely reduce the difficulties in optimization with neural volume rendering, thus enabling robust learning process. We also justify that the prior is the key factor that prevents collapse in the difficult sparse-view setting, we are able to obtain a rough but relatively complete geometric surface.

An intuitive solution is first to train an SDF Network for on-surface points and then use it as a part of NeuS directly. This is like pre-training the SDF function by on-surface points and then fine-tuning it with 2D images. However, fitting an SDF field to a sparse point cloud is challenging, due to the geometry complexity and the closed surface assumption. Even though our reconstruction targets are all closed surfaces, in sparse view, the on-surface points computed by the SfM method are only part of the surface of the objects. It means that enforcing the closed surface assumption leads to the overfitting of the geometric field.

To resolve this issue, the unsigned distance function (UDF) is considered to fit the global geometric field of the on-surface points. It is flexible and can cope with open surfaces. We train a UDF network $f_\theta$ to fit the surface points to obtain a complete and continuous geometric field. With the geometric field, we treat it as a coarse global prior that can stabilize the geometry optimizing with neural renderings.

Since the ground truth UDF values are not provided, we follow the CAP-UDF \cite{capudf} to train a network $f_\theta$ within a moving operation. We randomly sample a set of query locations $Q = \{\mathbf{q}_i,i\in[1,M]\}$ around given on-surface points $P$. Then we move the point $\mathbf{q}_i$ against the direction of the gradient at $\mathbf{\mathbf{q}}_i$ with a stride of UDF value $f_{\theta}(\mathbf{q}_i)$. Since the gradient points to the steepest uphill direction, the moving operation aims to find a path to pull point $\mathbf{p}_i$ onto the surface $S$. The operation can be formulated as:
\begin{equation}
    \label{eq:move}
    \mathbf{z}_i = \mathbf{q}_i - f_{\theta}(\mathbf{q}_i) \times \nabla f_{\theta}(\mathbf{q}_i)/||\nabla f_{\theta}(\mathbf{q}_i)||_2.
\end{equation}
Here, $\mathbf{z}_i$ is the location after the moving operation. The moving operation is differentiable in both the unsigned
distance value and the gradient, which allows us to optimize them simultaneously during training. 

For a well-learned network $f_{\theta}$, the moved points should be on the surface, which can be used as our learning objective. Hence, the Chamfer Distance between the moved points and the on-surface points can be used as $\mathcal{L}_{udf}$ to optimize our UDF network $f_{\theta}$:
\begin{equation}
\mathcal{L}_{udf} = {\rm CD}(Z,Q).
\label{udfloss}
\end{equation}

With a trained UDF network $f_\theta$, we attempt to incorporate the knowledge of the continuous geometry field into neural rendering. For the effect of network $f_\theta$, query points closer to the surface are more accurate. Therefore, we design a cut-off threshold $\epsilon$ for network $f_\theta$ to supervise the geometric field near the surface. The global geometry alignment is given by
\begin{equation}
\mathcal{L}_{global} = \frac{1}{|N|}\sum_{\mathbf{n}_i\in N} |f(\mathbf{n}_i)|\left(1-\frac{\max(f_{\theta}(\mathbf{n}_i)-\epsilon,0)}{f_{\theta}(\mathbf{n}_i)-\epsilon}\right),
\label{globalloss}
\end{equation}
where $N$ and $\epsilon$ are the discrete points in ray rendering and on-surface alignment threshold value, respectively. 

\begin{table*}[!ht]
    \centering
    \resizebox{\textwidth}{!}{
        \begin{tabular}{l|*{15}{c}|c}
            \toprule
            Scan ID & 24 & 37 & 40 & 55 & 63 & 65 & 69 & 83 & 97 & 105 & 106 & 110 & 114 & 118 & 122 & Mean \\
            \midrule
            \midrule
            \multicolumn{17}{c}{\emph{Little-overlap (PixelNeRF Setting)} } \\
            \midrule
            {COLMAP \cite{colmap}} & 2.88 & 3.47 & \textbf{1.74} & 2.16 & 2.63 & 3.27 & 2.78 & 3.63 & 3.24 & 3.49 & 2.46 & 1.24 & 1.59 & 2.72 & 1.87 & 2.61\\
            \cmidrule(l{0.7em}r{0.7em}){1-17}
            {SparseNeuS$_{ft}$ \cite{sparseneus}} & 4.81 & 5.56 & 5.81 & 2.68 & 3.30 & 3.88 & 2.39 & 2.91 & 3.08 & 2.33 & 2.64 & 3.12 & 1.74 & 3.55 & 2.31 & 3.34  \\
            {VolRecon \cite{volrecon}} & 3.05 & 4.45 & 3.36 & 3.09 & 2.78 & 3.68 & 3.01 & 2.87 & 3.07 & 2.55 & 3.07 & 2.77 & 1.59 & 3.44 & 2.51 & 3.02 \\
            \cmidrule(l{0.7em}r{0.7em}){1-17}
            NeuS \cite{neus} & 4.11 & 5.40 & 5.10 & 3.47 & 2.68 & 2.01 & 4.52 & 8.59 & 5.09 & 9.42 & 2.20 & 4.84 & 0.49 & 2.04 & 4.20 & 4.28 \\
            VolSDF \cite{volsdf} & 4.07 & 4.87 & 3.75 & 2.61 & 5.37 & 4.97 & 6.88 & 3.33 & 5.57 & 2.34 & 3.15 & 5.07 & 1.20 & 5.28 & 5.41 & 4.26  \\
            MonoSDF \cite{monosdf} & 3.47 & 3.61 & 2.10 & 1.05 & 2.37 & \textbf{1.38} & 1.41 & 1.85 & 1.74 & 1.10 & 1.46 & 2.28 & 1.25 & 1.44 & 1.45 & 1.86 \\
            \cmidrule(l{0.7em}r{0.7em}){1-17}
            Ours & \textbf{1.35} & \textbf{3.25} & 2.50 & \textbf{0.80} & \textbf{1.21} & 2.35 & \textbf{0.77} & \textbf{1.19} & \textbf{1.20} & \textbf{1.05} & \textbf{1.05} & \textbf{1.21} & \textbf{0.41} & \textbf{0.80} & \textbf{1.08} & \textbf{1.35} \\
            \midrule
            \midrule
            \multicolumn{17}{c}{\emph{Large-overlap (SparseNeuS Setting)}} \\
            \midrule
            {COLMAP \cite{colmap}} & 0.90 & 2.89 & 1.63 & 1.08 & 2.18 & 1.94 & 1.61 & 1.30 & 2.34 & 1.28 & 1.10 & 1.42 & 0.76 & 1.17 & 1.14 & 1.52 \\
            \cmidrule(l{0.7em}r{0.7em}){1-17}
            {SparseNeuS$_{ft}$ \cite{sparseneus}} & 2.17 & 3.29 & 2.74 & 1.67 & 2.69 & 2.42 & 1.58 & 1.86 & 1.94 & 1.35 & 1.50 & 1.45 & 0.98 & 1.86 & 1.87 & 1.96 \\
            {VolRecon \cite{volrecon}} & 1.20 & 2.59 & 1.56 & 1.08 & 1.43 & 1.92 & 1.11 & 1.48 & 1.42 & 1.05 & 1.19 & 1.38 & 0.74 & 1.23 & 1.27 & 1.38 \\
            \cmidrule(l{0.7em}r{0.7em}){1-17}
            NeuS \cite{neus} & 4.57 & 4.49 & 3.97 & 4.32 & 4.63 & 1.95 & 4.68 & 3.83 & 4.15 & 2.50 & 1.52 & 6.47 & 1.26 & 5.57 & 6.11 & 4.00  \\
            VolSDF \cite{volsdf} & 4.03 & 4.21 & 6.12 & 0.91 & 8.24 & 1.73 & 2.74 & 1.82 & 5.14 & 3.09 & 2.08 & 4.81 & 0.60 & 3.51 & 2.18 & 3.41 \\
            MonoSDF \cite{monosdf} & 2.85 & 3.91 & 2.26 & 1.22 & 3.37 & 1.95 & 1.95 & 5.53 & 5.77 & 1.10 & 5.99 & 2.28 & 0.65 & 2.65 & 2.44 & 2.93 \\
            \cmidrule(l{0.7em}r{0.7em}){1-17}
            Ours & \textbf{0.78} & \textbf{2.35} & \textbf{1.55} & \textbf{0.75} & \textbf{1.04} & \textbf{1.68} & \textbf{0.60} & \textbf{1.14} & \textbf{0.98} & \textbf{0.70} & \textbf{0.74} & \textbf{0.49} & \textbf{0.39} & \textbf{0.75} & \textbf{0.86} & \textbf{0.99} \\
            \bottomrule
        \end{tabular}
    }
    \caption{The quantitive comparison results of Chamfer Distances (CD$\downarrow$) on DTU dataset.}
    \label{tab:cd_results}
\end{table*}

\subsection{On-Surface Local Geometry Refinement}

Geometric field alignment can constrain the shape and ensure the integrity of the reconstructed object. However, the learned geometric field is coarse and can not hone the details of the reconstruction. The reason is that the sparse input provides less surface feature information, this may lead to severe noise. Therefore, a local-level optimization is needed.  Our insight comes from the traditional MVS methods \cite{mvsnet,surfacenet} where the correctness of a surface point estimation is guaranteed by the consistency of its corresponding feature between different views. We justify that on-surface points obtained by SfM methods also follow this assumption. Inspired by that, we render images with given poses and a novel pose. By supervising the projection features of the on-surface points, geometry and color fields are optimized simultaneously, which can be expressed by  
\begin{equation}
\mathcal{L}_{local} = \frac{1}{|P||I|}\sum_{\mathbf{p}_i\in P}\sum_{\mathcal{H}_j\in \mathcal{H}}\left \Vert f_{\phi}(\mathcal{H}_{j}(\mathbf{p}_i))-f_{\phi}(\mathcal{H}_{0}(\mathbf{p}_i))\right \Vert,
\label{featureloss1}
\end{equation}
where $P$ is on-surface point cloud obtained by SfM method, $f_{\phi}$ is a geometry feature extraction network, $I$ and $\mathcal{H}$ are input views and the transformation matrices respectively. 

\begin{figure}[!ht]
    \centering
    \includegraphics[width=0.47\textwidth]{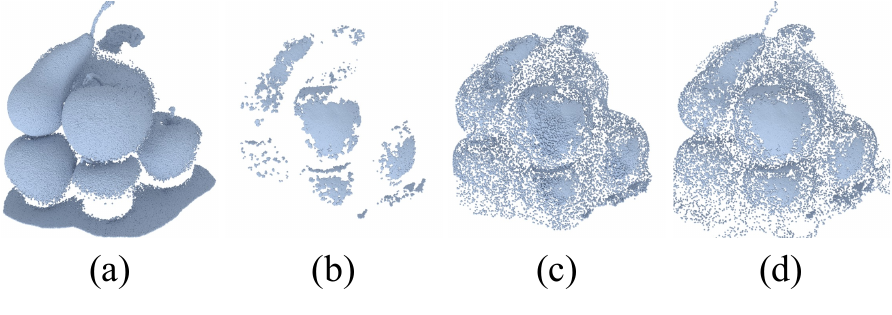}
    \caption{(a) Ground truth on-surface points; (b) On-surface points generated with SfM; (c) Pseudo and SfM on-surface points at early stages of training; (d) Pseudo and SfM on-surface points at the end of training process. The pseudo on-surface points are optimized to potential real surface.}
    \label{fig:points}
\end{figure}
However, initial points obtained by SfM method are limited in quantity and unevenly distributed. The points are barely concentrated on positions of relatively poor visibility, which leads to biased optimization spatially. Hence, during the ray rendering process, we obtain some points by calculating the ray-surface intersection points $P'$. We name them pseudo on-surface points since they are acquired from the implicit surface to be optimized. The specific point $\mathbf{p'}$ passed through by a ray $r$ during neural rendering is denoted as 
$\mathbf{p'} = \{\mathbf{r}(t^*)| f(\mathbf{r}(t^*))=0\}$. We first find $t_i$ by solving $f(\mathbf{r}(t_i))f(\mathbf{r}(t_{i+1}))<0$. And $t^*$ can be calculated as:
\begin{equation}
t^* = \frac{f(\mathbf{r}(t_i))t_{i+1}-f(\mathbf{r}(t_{i+1}))t_i}{f(\mathbf{r}(t_i))-f(\mathbf{r}(t_{i+1}))}.
\end{equation}

Since the computation of the pseudo on-surface points is differentiable, it is reasonable to optimize the surface by considering them within the feature projection alignment loss. The implicit surface together with 3D coordinates of the pseudo on-surface points themselves are optimized along the training process, as we can see in Figure \ref{fig:points}. 
Therefore, we add differentiable pseudo surface points $P'$ to surface points $P$ during the training process to keep the projected features consistent between seen and unseen views:
\begin{equation}
P\leftarrow P+P'.
\end{equation}

\begin{figure*}[!t]
  \centering
  \includegraphics[width=\textwidth]{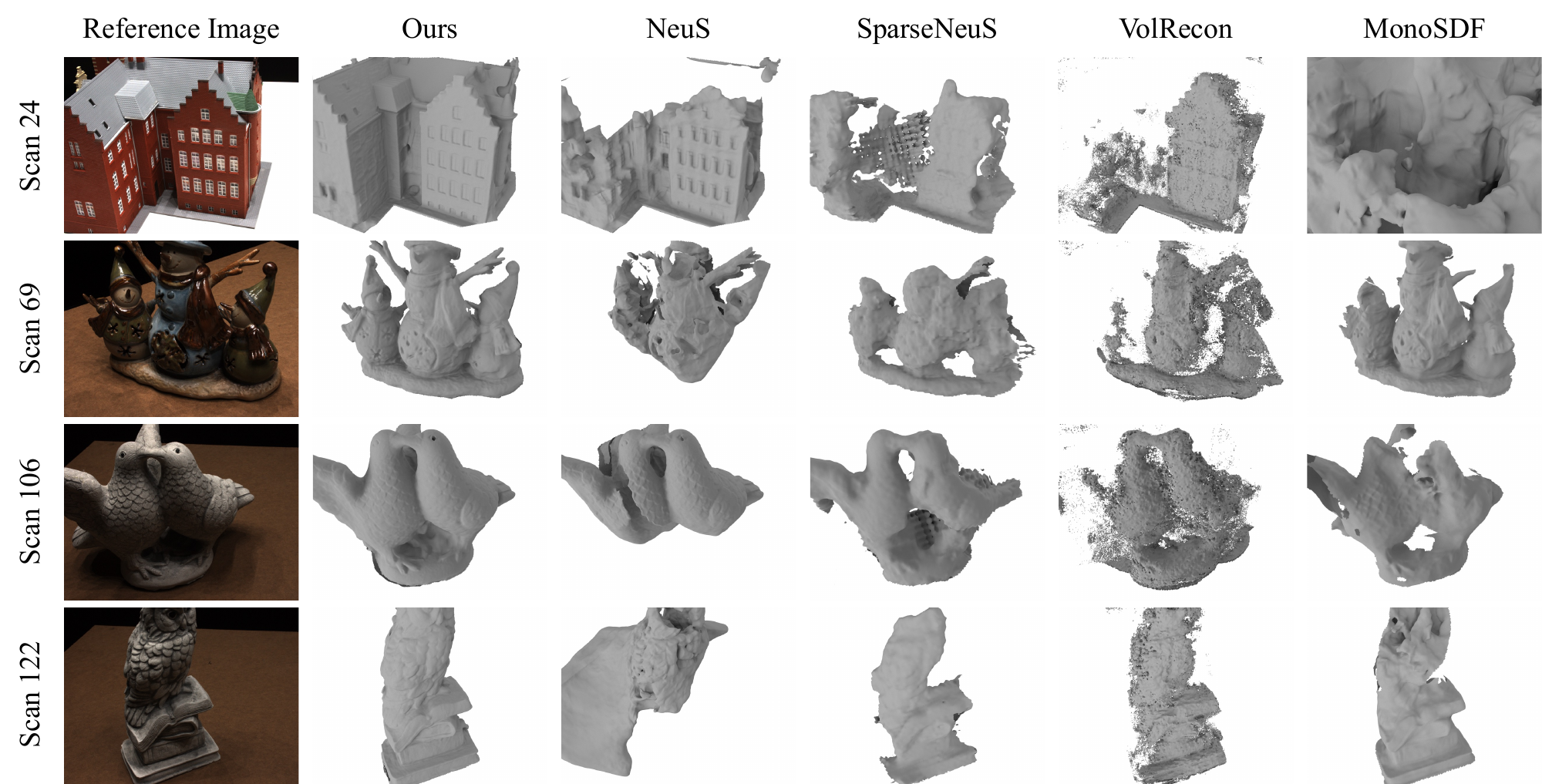}
  \caption{Visual comparisons of surface reconstruction results on the little-overlap sparse setting of DTU dataset. (*NeuS cannot generate valid mesh for scan 122 with the generic 3 views. We take an additional view for training on this scan with NeuS for visual comparison.)}
  \label{fig:little-overlap-results}
\end{figure*}

\subsection{Loss Functions}

The overall loss functions are:
\begin{equation}
\mathcal{L} =  \mathcal{L}_{color} +\lambda_{1} \mathcal{L}_{global}  +\lambda_{2} \mathcal{L}_{local} +\lambda_{3} \mathcal{L}_{eik}+\lambda_{4} \mathcal{L}_{reg},
\end{equation}
where $\mathcal{L}_{global}$ and $\mathcal{L}_{local}$ are the on-surface global geometry alignment loss
and local refinement loss defined above.

$\mathcal{L}_{color}$ is the difference between the rendered and ground-truth pixel colors:
\begin{equation}
\mathcal{L}_{color} = \frac{1}{\mathcal{R}}\sum_{\mathbf{r}\in{\mathcal{R}}}{||C(\mathbf{r})-\hat{C}(\mathbf{r})||}.
\end{equation}

As with NeuS \cite{neus}, an Eikonal term \cite{IGR} on the random sample points $\mathcal{Y}$ to regularize SDF of $f(x)$ is introduced: 
\begin{equation}
\mathcal{L}_{eik} = \frac{1}{|\mathcal{Y}|}\sum_{\mathbf{x}\in{\mathcal{Y}}}{(||\nabla f(\mathbf{x})||-1)^2}.
\end{equation}

Similar to Geo-NeuS \cite{geoneus}, AutoRecon \cite{autorecon}. We adopt $\mathcal{L}_{reg}$ as supervision with a zero-level set. However, given sparse views, the points we obtain by SfM are sparse and precious. We do not supervise rendered depth with real depth for each view. Alternatively, We supervise the SDF values of all points in 3D space in each iteration of training:
\begin{equation}
\mathcal{L}_{reg} = \frac{1}{P}\sum_{\mathbf{p}_i\in{P}}{||f(\mathbf{p}_i)||}.
\end{equation}

\section{Experiments and Analysis}
We conduct abundant surface reconstruction experiments on several generic public MVS datasets \cite{dtu,blendedmvs} from sparse views. We compare our results with some recently presented neural implicit surface reconstruction methods, including the previous state-of-the-art sparse views specified methods. Then we give an ablation study of our approach.

\begin{figure*}[!t]
  \centering
  \includegraphics[width=\textwidth]{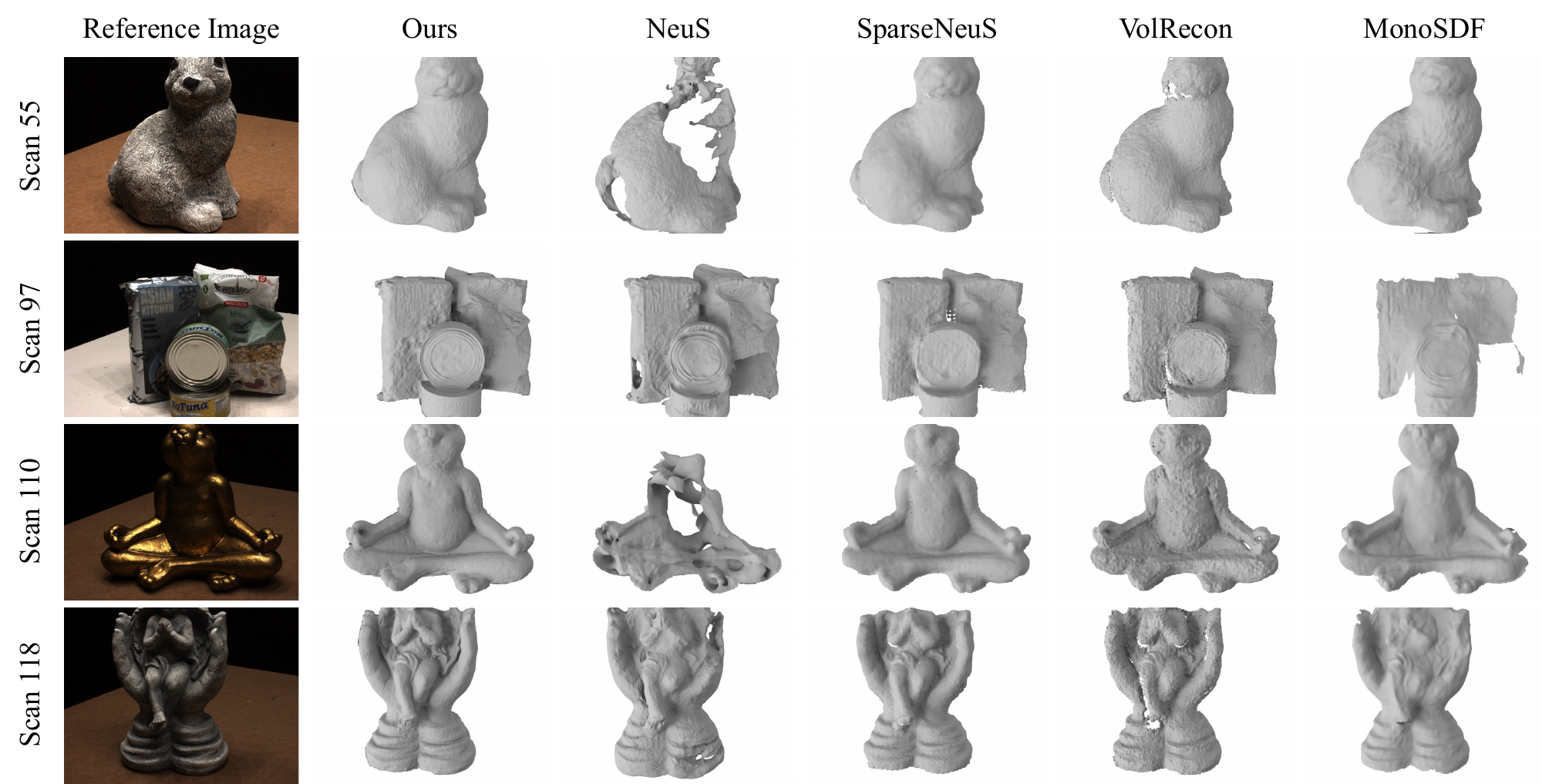}
  \caption{Visual comparisons of surface reconstruction results on the large-overlap sparse setting of DTU dataset.}
  \label{fig:large-overlap-results}
\end{figure*}

\subsection{Experimental Settings}

\subsubsection{Datasets.}
Previous neural sparse view reconstruction approaches normally select 3 proper input views from each scan of the DTU dataset \cite{dtu}, which contains from 49 to 64 images at a resolution of $1200 \times 1600$ for each object scan with known camera intrinsics and poses, to evaluate the performance of the models. We notice that different approaches differ in the choice of concrete input views. SparseNeus \cite{sparseneus} and VolRecon \cite{volrecon} take views 23, 24 and 33 of each scan as one of the test sets for three-view reconstruction. We name it \emph{large-overlap} setting because the distribution of the selected views is concentrated and the visibility overlap between the pics is relatively large. While MonoSDF \cite{monosdf} follows PixelNeRF \cite{monosdf}, taking views 22, 25 and 28 of each scan as sparse-view setting, which we name \emph{little-overlap} setting because of the scattered view distribution. To evaluate our approach and the baselines comprehensively and fairly, we conduct experiments on both two settings on 15 scans commonly used for evaluation.

Besides, we also employ the BlendedMVS dataset \cite{blendedmvs} to estimate the versatility of our approach. We randomly select 3 views from each scene as input and conduct the evaluation on 8 challenging scenes at a resolution of 768 $\times$ 576.

\subsubsection{Baselines.}

We compare our approach with various types of surface reconstruction methods on adopted datasets. \textbf{a.} COLMAP \cite{colmap}: A widely used classical SfM framework, which is also the pro-precessing approach we employ in our pipeline. \textbf{b.} Generalizable neural implicit surface reconstruction methods, including SparseNeuS${_{ft}}$ \cite{sparseneus} and VolRecon \cite{volrecon}. $ft$ indicates that we do fine-tuning on every single scene before we test the model. \textbf{c.} Per-scene optimization methods, including NeuS \cite{neus}, VolSDF \cite{volsdf} and MonoSDF \cite{monosdf}. We adjust the experiment settings for specific baselines to maximize their performance.

\subsubsection{Implementation details.}
We use naive COLMAP \cite{colmap} to obtain the coarsely estimated point clouds of the test scenes with ground truth poses as inputs. We implement SDF representation model and neural radiance field based on NeuS \cite{neus} baseline and adopt similar network architecture as CAP-UDF \cite{capudf} to learn UDF network $f_\theta$. To achieve better local geometry refinement, we use Vis-MVSNet \cite{vismvsnet} as the feature extraction network $f_\phi$. 

For a training procedure of a single scene, we sample 512 rays per batch and train the model for 300k iterations on an NVIDIA RTX3090 GPU.

\begin{figure}[!ht]
    \centering
    \includegraphics[width=0.45\textwidth]{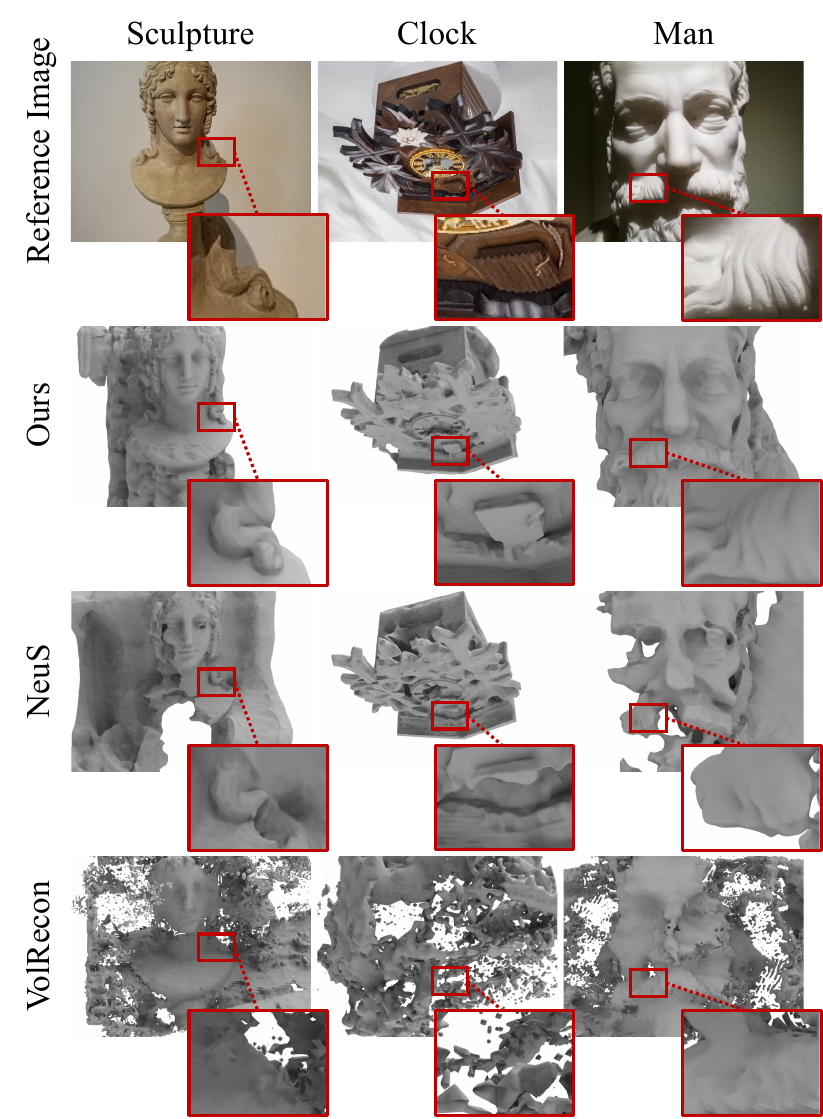}
    \caption{Visual comparison of surface reconstruction results on BlendedMVS dataset.}
    \label{fig:bmvs_results}
\end{figure}

\subsection{Comparisons}

\subsubsection{Sparse View Reconstruction on DTU.}
We conduct comparisons on both two DTU sparse settings without mask supervision. We measure the Chamfer Distances on DTU dataset in the same way as \cite{volrecon} to quantitatively evaluate the reconstruction quality, which is demonstrated in Table \ref{tab:cd_results}. Our approach achieves better performance on most of the scenes in little-overlap setting and all of the scenes in large-overlap setting, outperforming the compared baselines including previous state-of-the-art methods.

We present visualizations for both types of DTU sparse-settings. Figure \ref{fig:little-overlap-results} shows the reconstruction results on little-overlap setting, while Figure \ref{fig:large-overlap-results} shows the reconstruction results of the approaches on large-overlap setting. It is challenging for most of the compared methods to obtain complete geometry when the distribution of input views is relatively concrete, while our approach not only captures enough global information to basically rebuild the correct coarse shape but also better restore the facial details of the objects.

\subsubsection{Sparse View Reconstruction on BlendedMVS.}
To evaluate the generalization ability of our approach on different datasets, we perform an evaluation on BlendedMVS dataset. We conduct reconstruction tests on 8 representative scenes, from each of which 3 views are randomly selected together with the corresponding camera poses. Some of the reconstruction results are visualized in Figure \ref{fig:bmvs_results}. It shows that our approach could obtain both better global shapes and finer geometric details.

\subsection{Ablation and Analysis}
\begin{figure}[!ht]
    \centering
    \includegraphics[width=0.4\textwidth]{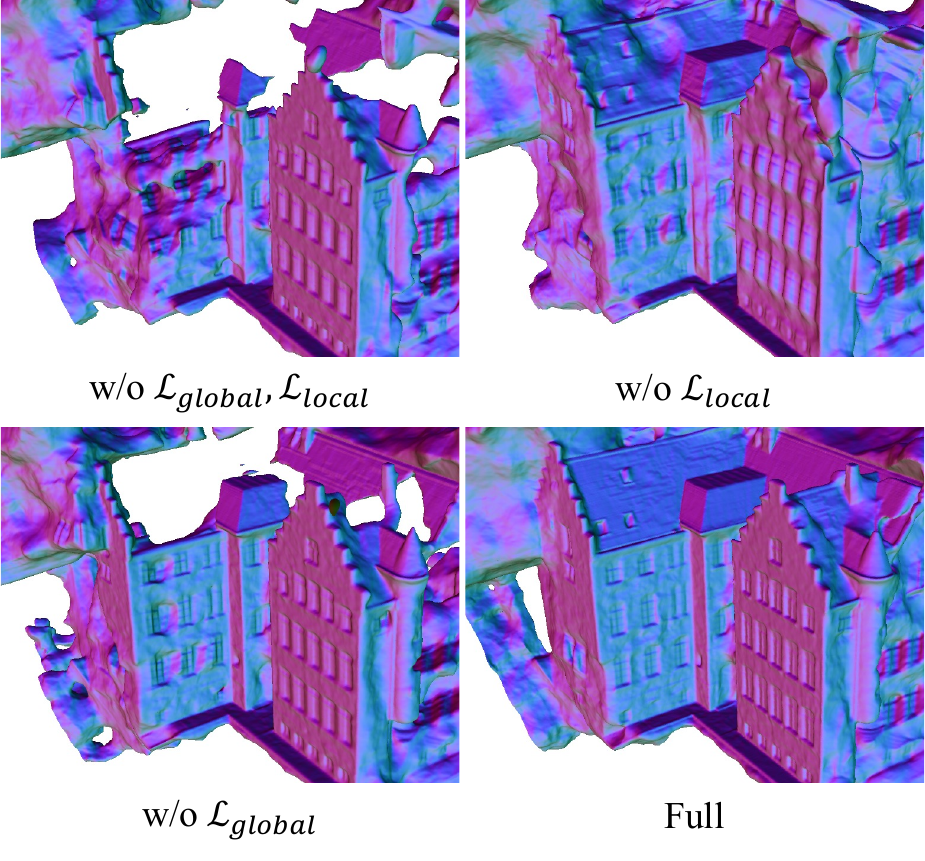}
    \caption{Comparison for reconstructed normal maps of ablation results on DTU scan 24. }
    \label{fig:ablation_results}
\end{figure}
On-surface global geometric loss $\mathcal{L}_{global}$ and local geometric loss $\mathcal{L}_{local}$ serve as two main components of our reconstruction approach. To better evaluate the effectiveness of these supervisions, we conduct an ablation study. We test our models on the little-overlap DTU sparse setting because this would better reflect the performance of the methods. We separately evaluate the model without $\mathcal{L}_{global}$, the model without $\mathcal{L}_{local}$, and the model without both two losses on all 15 scenes. The mean Chamfer Distances are demonstrated in Table \ref{tab:ablation}.

\setlength\tabcolsep{20pt}
\begin{table}[t!]
\centering
\begin{tabular}{ c c | c }
\toprule
$\mathcal{L}_{global}$ & $\mathcal{L}_{local}$ & Mean CD$\downarrow$ \\
\midrule
\scalebox{0.85}[1]{$\times$} & \scalebox{0.85}[1]{$\times$} & 2.46 \\
\scalebox{0.85}[1]{$\times$} & \raisebox{0.6ex}{\scalebox{0.7}{$\sqrt{}$}} & 1.96 \\
\raisebox{0.6ex}{\scalebox{0.7}{$\sqrt{}$}} & \scalebox{0.85}[1]{$\times$} & 1.67 \\
\raisebox{0.6ex}{\scalebox{0.7}{$\sqrt{}$}} & \raisebox{0.6ex}{\scalebox{0.7}{$\sqrt{}$}} & \textbf{1.35} \\
\bottomrule
\end{tabular}

\caption{Reconstruction results comparison of mean Chamfer Distance on little-overlap sparse input subset of DTU dataset by variants of our approach.}
\label{tab:ablation}
\end{table}

Although even the model without $\mathcal{L}_{global}$ and $\mathcal{L}_{local}$ still outperforms NeuS baseline, global geometric prior and local feature projection boost the performance to a huge extent. To point out the concrete contributions of these components more intuitively, we give a visualization of the reconstructed normal maps of these ablation models on a single scene in Figure \ref{fig:ablation_results}. As we can see from the comparison, the rarity of input views leads to the dislocation of some local structures. The participation of $\mathcal{L}_{local}$ alleviates the error. The sparsity of view distribution also introduces a new problem: some spatially continuous parts are incomplete out of the hardness to distinguish foreground and background. Global UDF prior significantly improves the integrity, even when the point clouds obtained by COLMAP are fragmented.

\section{Conclusion}

In this paper, we proposed NeuSurf, a novel sparse view surface reconstruction method with on-surface priors. To obtain a rough and complete
geometric surface, we train a UDF network to learn the on-surface geometry field and leverage it as global geometry alignment. Then we optimize the feature consistency as local geometry refinement loss to reconstruct detailed surfaces. Our method does not require large-scale training and is robust in various sparse settings. Our method achieves state-of-the-art performance on the DTU dataset in both large-overlap and little-overlap settings. Additionally, we conduct qualitative experiments on the BlendedMVS dataset in different sparse settings and find significant improvement over previous methods.

\section{Acknowledgments}
This work was supported by the Program of Science and Technology Plan of Beijing (Z231100001723014).
\bibliography{aaai24}
\clearpage


\section*{Supplementary material}

# Supplementary Materials for "NeuSurf: On-Surface Priors for Neural Surface Reconstruction from Sparse Input Views"

In this supplementary document, we first discuss architectural and implementation details in Section **A**. Next, we provide additional ablation studies and report additional quantitative and qualitative results in Section **B** and Section **C**. Following that, in order to further interpret and analyze the reconstruction results of COLMAP, we have conducted both qualitative and quantitative comparisons in Section **D**. Then, We summarize all relevant work according to the sparse setting and training method in Section **E**. Finally, we discuss the potential social negative impact of this work in Section **F**.

## A. Details of Architectural and Implementation

**Architectures**

In the main paper, we propose a novel framework NeuSurf for surface reconstruction from sparse view images. overall, NeuSurf shares a similar base structure to NeuS (Wang et al. 2021), but introduce two main constraints for sparse view reconstruction. Figure 1 is a comparison of our framework and NeuS. The key idea is to use the global geometric fields obtained from the surface points to help learn rough and continuous geometry, and optimize the local feature consistency of on-surface points to help learn fine geometry.

**On-Surface Global Geometry Alignment**

Inspired by CAP-UDF, we first generate a bunch of query points $Q$ near the surface point cloud, and learn a UDF Network $f_\theta$ from the point cloud without ground truth UDF values by moving the query points $Q$ to the surface $S$. The operation of moving the query point is differentiable. By optimizing the distribution difference between the moved query point and the real surface point, we can optimize the expression of the entire UDF Network, which is a solid field learning method. According to our main text, the move operation can be expressed as

$$\mathbf{z}_i = \mathbf{q}_i - f_\theta(\mathbf{q}_i) \times \nabla f_\theta(\mathbf{q}_i) / ||\nabla f_\theta(\mathbf{q}_i)||_2. \quad (1)$$

We use chamfer distance as the learning goal of UDF Network

$$\mathcal{L}_{udf} = \mathrm{CD}(Z, Q). \quad (2)$$

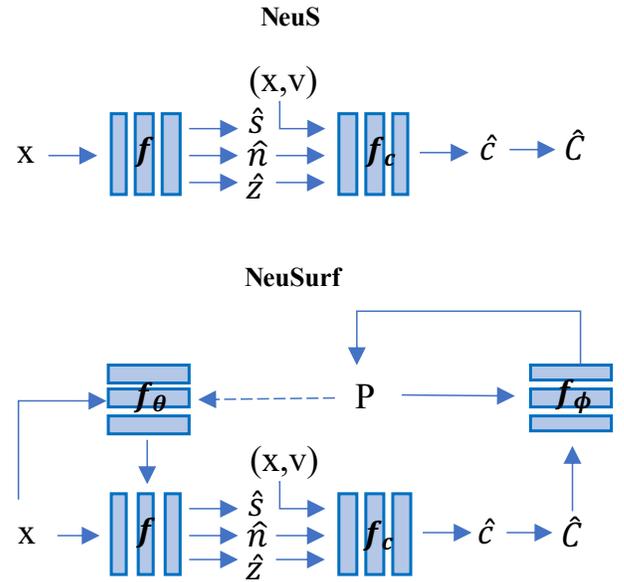

Figure 1: NeuS and NeuSurf.

According to the definition of chamfer distance (Barrow et al. 1977), CD can be expressed as

$$\mathrm{CD}(Z, Q) = \frac{1}{M} \sum_{i \in [1,M]} \min_{j \in [1,N]} ||\mathbf{z}_i - \mathbf{p}_j||_2 \\ + \frac{1}{N} \sum_{j \in [1,N]} \min_{i \in [1,M]} ||\mathbf{p}_j - \mathbf{z}_i||_2. \quad (3)$$

Using the chamfer distance loss function to optimize $f_\theta$, we can get a continuous and complete field function $f_\theta$ instead of discrete points $P$ to complete global the geometric alignment.

The global geometry alignment is given in the main paper

$$\mathcal{L}_{global} = \frac{1}{|N|} \sum_{\mathbf{n}_i \in N} |f(\mathbf{n}_i)| \left(1 - \frac{\max(f_\theta(\mathbf{n}_i) - \epsilon, 0)}{f_\theta(\mathbf{n}_i) - \epsilon}\right), \quad (4)$$

## On-Surface Local Geometry Refinement

In the task of few-shot NeRF (Yu et al. 2021), many recent methods (Jain, Tancik, and Abbeel 2021; Kim, Seo, and Han 2022) optimize the field with rendering results from unseen viewpoints. But there is no ground truth for color from unseen views, and only some semantic-based methods can be used to optimize the color field and density field.

In the paper, we impose on-surface local geometry refinement to improve the details of surfaces. Our goal is to ensure that the points on the surface meet the conditional assumptions of multi-vie consistency between seen and unseen views. We render images with given poses and a novel pose. By supervising the projection features of the on-surface points, geometry and color fields are optimized simultaneously, which can be mathematically expressed by

$$\mathcal{L}_{local} = \frac{1}{|P||I|} \sum_{\mathbf{p}_i \in P} \sum_{\mathcal{H}_j \in \mathcal{H}} \|f_\phi(\mathcal{H}_j(\mathbf{p}_i)) - f_\phi(\mathcal{H}_0(\mathbf{p}_i))\|, \quad (5)$$

where $\mathcal{H}_j(\mathbf{p}_i)$ can be calculated as:

$$\mathcal{H}_j(p_i) = \mathbf{K_j}(\mathbf{R_j p}_i + \mathbf{t_j}). \quad (6)$$

$\mathbf{K_j}$ and $[\mathbf{R_j}, \mathbf{t_j}]$ are the corresponding camera intrinsic and extrinsic parameters.

In order to further strengthen our prior assumptions, we continuously calculate pseudo on-surface points in the process of neural rendering, and add them into the on-surface point set to jointly optimize our geometric field.

## Implementation Details

The overall loss functions are:

$$\mathcal{L} = \mathcal{L}_{color} + \lambda_1 \mathcal{L}_{global} + \lambda_2 \mathcal{L}_{local} + \lambda_3 \mathcal{L}_{eik} + \lambda_4 \mathcal{L}_{reg}, \quad (7)$$

To obtain faithful reconstruction results, the hyperparameters $\lambda_1, \lambda_2, \lambda_3, \lambda_4$ are set to $1.0, 0.1, 1.0, 0.1, 1.0$.

For the obtained on-surface points, to get a credible on-surface global geometry alignment, we adopt a UDF network as $f_\theta$, similar to CAP-UDF, which contains 8 256-node layers of MLP and a skip connection in the fourth layer. Referring to most UDF network designs, before the network outputs the result, we take the absolute value of the output value to ensure that the output result is a UDF value. In addition, we use the feature extractor in Vis-MVSNet as $f_\phi$ to extract the features of the seen and unseen image to ensure the feature consistency of the initial and pseudo on-surface points for on-surface local geometry refinement.

## Sparse Setting Details

There are currently two popular sparse-settings for DTU dataset, we have summarized and named these two, pointed out their differences, and then visualized their distribution differences.

SparseNeuS (Long et al. 2022) and VolRecon (Ren et al. 2023) take views 23, 24 and 33 for sparse view reconstruction. We name it *large-overlap* setting because the pose distribution of the selected views is concentrated and the visibility overlap between the pictures is large.

MonoSDF (Yu et al. 2022) follows PixelNeRF (Yu et al. 2022), taking views 22, 25 and 28 as sparse view setting, which we name *little-overlap* setting because of the scattered view distribution and little overlap between images.

We visualize the camera poses and orientations for two different sparse settings in Figure 2. As we describe, the distribution of large-overlap camera poses in blue is more tightly distributed, while the distribution of poses of orange little-overlap cameras is more discrete. Because the datasets are taken around the circle, their camera orientations are all pointing towards the object's surface.

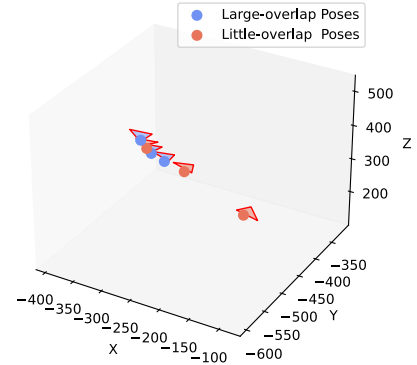

Figure 2: Sparse Settings Poses Visualization.

## B. Additional Ablation Study

As we mentioned in the main text, SDF field, as an acceptable representation of surface point geometric continuous field, can also be used for global geometric alignment. However, the geometry complexity and the closed surface assumption limit its use in such sparse view conditions. To corroborate our statement, we conduct ablation experiments with SDF field priors. Here, we use the method of Neural-pull(Ma et al. 2021) to train an SDF network $f_\theta$. Similar to the supervision of UDF, we directly use the value of SDF to constrain the geometric function $f$ in neural rendering. The results are shown in Table 3

| SDF Prior | UDF Prior | Mean CD↓ |
|:---:|:---:|:---:|
| √ | × | 2.49 |
| × | √ | **1.35** |

Table 1: Comparison of reconstruction with different global geometric priors on DTU little-overlap setting.

## C. More Quantitative and Qualitative Results

Geo-NeuS (Fu et al. 2022) is the SOTA neural-rendering method in dense view reconstruction on DTU (Jensen et al. 2014) dataset. To demonstrate the priority of our method,

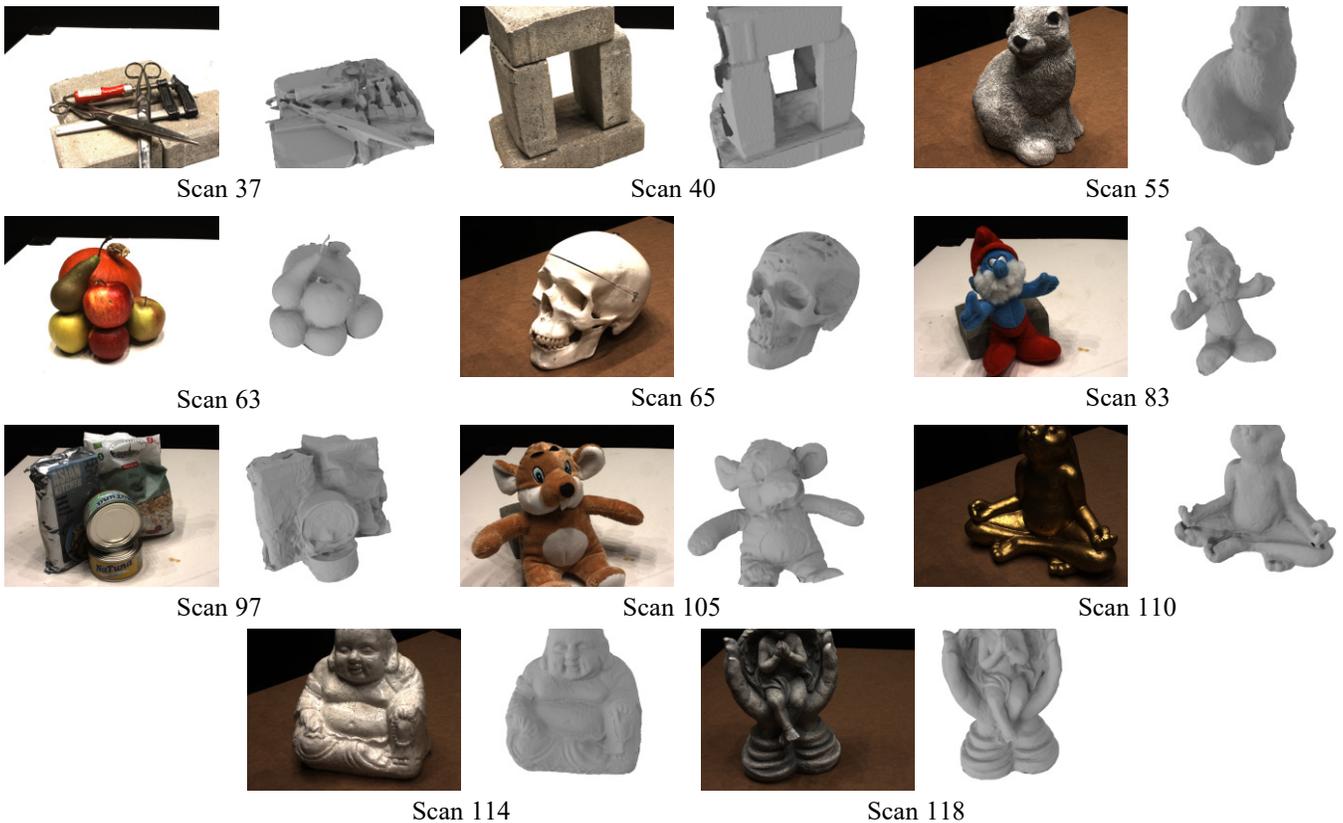

Figure 3: The remaining visual results of surface reconstruction on the little-overlap sparse setting of DTU dataset.

we further apply Geo-NeuS to sparse view reconstruction and compare it with our method, as shown in Table 2.

It can be seen that there is a gap when the SOTA reconstruction method of dense views is directly used for sparse view reconstruction. Geo-NeuS uses the projection depth of discrete points to optimize the geometry, which makes sense under the experimental setting of dense views, but the point cloud of the sparse view is more sparsely distributed, discontinuous and incomplete optimization may lead to localized geometry collapse. Whereas our method addresses this issue by learning the continuous geometric field of point clouds, which makes better use of the geometric priors of on-surface points.

In our paper, we show several representative scenes for visual comparison on DTU dataset. To demonstrate the integrity of the experiment, in this supplementary, we present qualitative results with DTU (Jensen et al. 2014) datasets in Figure 3 and Figure 4 for the remaining 11 scenes in little-overlap setting and large-overlap setting.

For the BlendedMVS dataset, we also show the visualization results of more scenes in Figure 6. According to the comparison, it can be found that our method can achieve the best results no matter what kind of sparse setting and dataset.

## D. Comparision with COLMAP Reconstruction

The results from COLMAP consist of point clouds rather than mesh surfaces. Similar to previous methods, we have included it solely for reference purposes. To enhance the clarity of our results, we utilized Poisson Reconstruction (Kazhdan and Hoppe 2013) and OnSurfacePrior (Baorui, Yu-Shen, and Zhizhong 2022) to generate mesh surfaces from the COLMAP points on DTU little-overlap setting. As shown in Figure 5, compared with the results of COLMAP points and surfaces, our results are more complete and detailed.

In order to further explore, we conduct quantitative comparisons for several non-NeuS related reconstruction methods (models not based on the NeuS backbone). The results are shown in Table 3. OnSurfacePrior is proposed to optimize reconstruction on sparse point clouds rather than incomplete points generated from sparse views, while NeuralUDF takes a similar pipeline for UDF field reconstruction without refinement on sparse-views scenes. Our method's reconstruction results outperform theirs.

## E. Concurrent work

We summarize all sparse view implicit reconstruction methods (including published articles and preprinted articles on arxiv). From what we mentioned in the main text, some

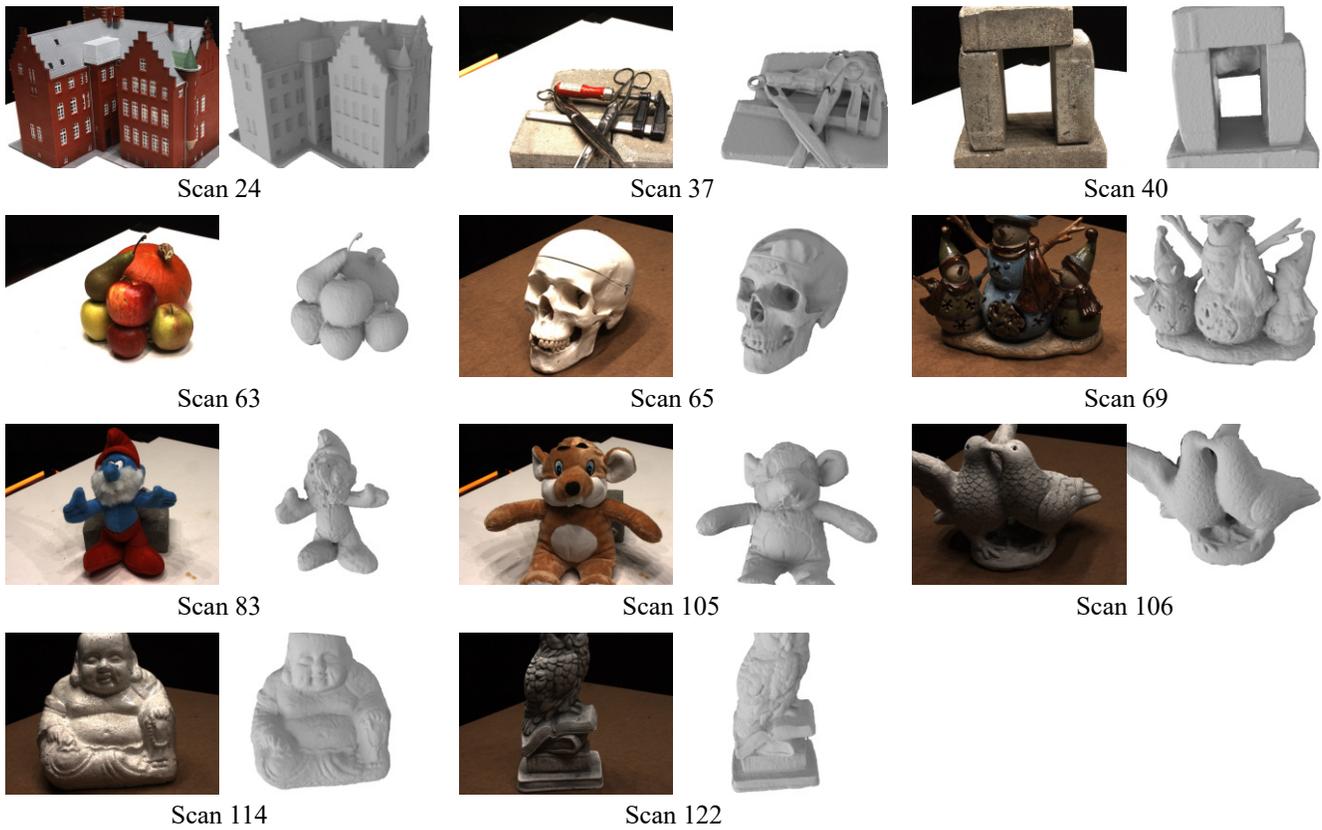

Figure 4: The remaining visual results of surface reconstruction on the large-overlap sparse setting of DTU dataset.

| Scan ID | 24 | 37 | 40 | 55 | 63 | 65 | 69 | 83 | 97 | 105 | 106 | 110 | 114 | 118 | 122 | Mean |
|---|---|---|---|---|---|---|---|---|---|---|---|---|---|---|---|---|
| *Little-overlap (PixelNeRF Setting)* | | | | | | | | | | | | | | | | |
| Geo-NeuS | 9.26 | 10.86 | * | 1.39 | 8.39 | 6.49 | 2.00 | 1.80 | 1.90 | 1.22 | 1.40 | **0.89** | 0.53 | 1.27 | 1.58 | 3.50 |
| Ours | **1.35** | **3.25** | **2.50** | **0.80** | **1.21** | **2.35** | **0.77** | **1.19** | **1.20** | **1.05** | **1.05** | 1.21 | **0.41** | **0.80** | **1.08** | **1.35** |
| *Large-overlap (SparseNeuS Setting)* | | | | | | | | | | | | | | | | |
| Geo-NeuS | 3.55 | 13.99 | 3.03 | 0.77 | 1.56 | 5.01 | 0.74 | 1.16 | 1.00 | **0.68** | 0.84 | 0.52 | 0.41 | 0.78 | 0.92 | 2.33 |
| Ours | **0.78** | **2.35** | **1.55** | **0.75** | **1.04** | **1.68** | **0.60** | **1.14** | **0.98** | 0.70 | **0.74** | **0.49** | **0.39** | **0.75** | **0.86** | **0.99** |

Table 2: The quantitive comparison results of Chamfer Distances (CD↓) on DTU dataset. (*Geo-NeuS failed to generate valid mesh on Scan 40 of little-overlap setting.)

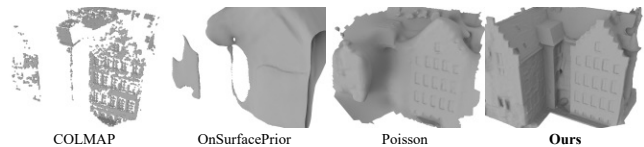

COLMAP  OnSurfacePrior  Poisson  **Ours**

Figure 5: Visual comparison of reconstruction on 3-view images from DTU.

| Method | NeuralUDF | OnSurfacePrior | Poisson | Ours |
|---|---|---|---|---|
| Mean CD↓ | 5.16 | 4.59 | 3.27 | **1.35** |

Table 3: More comparison on DTU little-overlap setting.

methods use three views with a large overlap to complete sparse view reconstruction.

- **SparseNeuS (Long et al. 2022) in ECCV 2022** is the first work to adopt a large-overlap setting for sparse view reconstruction. They present a novel neural surface reconstruction method, which trains for days on large amounts of data, then generalizes to new scenes to achieve better reconstruction results.

- **VolRecon (Ren et al. 2023) in CVPR 2023** follows the sparse-setting of SparseNeuS. They introduce a novel generalizable implicit reconstruction method with

Signed Ray Distance Function. Same as the SparseNeuS, they also require longer training time to learn a prior that generalizes well.

- **C2F2NeuS (Xu et al. 2023) in arXiv 2023** combines the multi-view stereo with neural signed distance function representations. This paper is still a preprint and not yet open source. It is the closest performance to our work in the large-overlap setting.

others use three views with a little overlap to complete sparse view reconstruction.

- **MonoSDF (Yu et al. 2022) in NeurIPS 2022** introduces depth and normal geometric priors for dense reconstruction, and also applies them to the reconstruction of sparse views. Unlike SparseNeuS, the sparse setting adopted by MonoSDF is consistent with PixelNeRF(Yu et al. 2021), which is also widely used in few-shot NeRF task.
- **S-VolSDF (Wu, Graikos, and Samaras 2023) in arXiv 2023** follows the little-overlap sparse setting. They use the MVS method to optimize the neural rendering process, and use neural rendering to guide the calculation results of MVS method. This paper is still a preprint and not yet open source.
- **DiViNet (Vora, Patil, and Zhang 2023) in arXiv 2023** for the case of little overlap, a neural template-based method is proposed for constrained geometric learning. The templates need training across different scenes and serve as anchors in new scenes. This paper is still a preprint and not yet open source.

Although these methods (published papers or unpublished preprints) have their own advantages under different settings, extensive comparative experiments prove that **our method outperforms all the above methods in both sparse settings.**

### F. The potential negative impact of this work

The ability to reconstruct an object's surface with only a few photos may allow the geometric information of the object to be obtained without authorization. Our reconstruction method requires hours of training time on high-performance GPU servers, which consumes energy and has a negative impact on global environmental change.

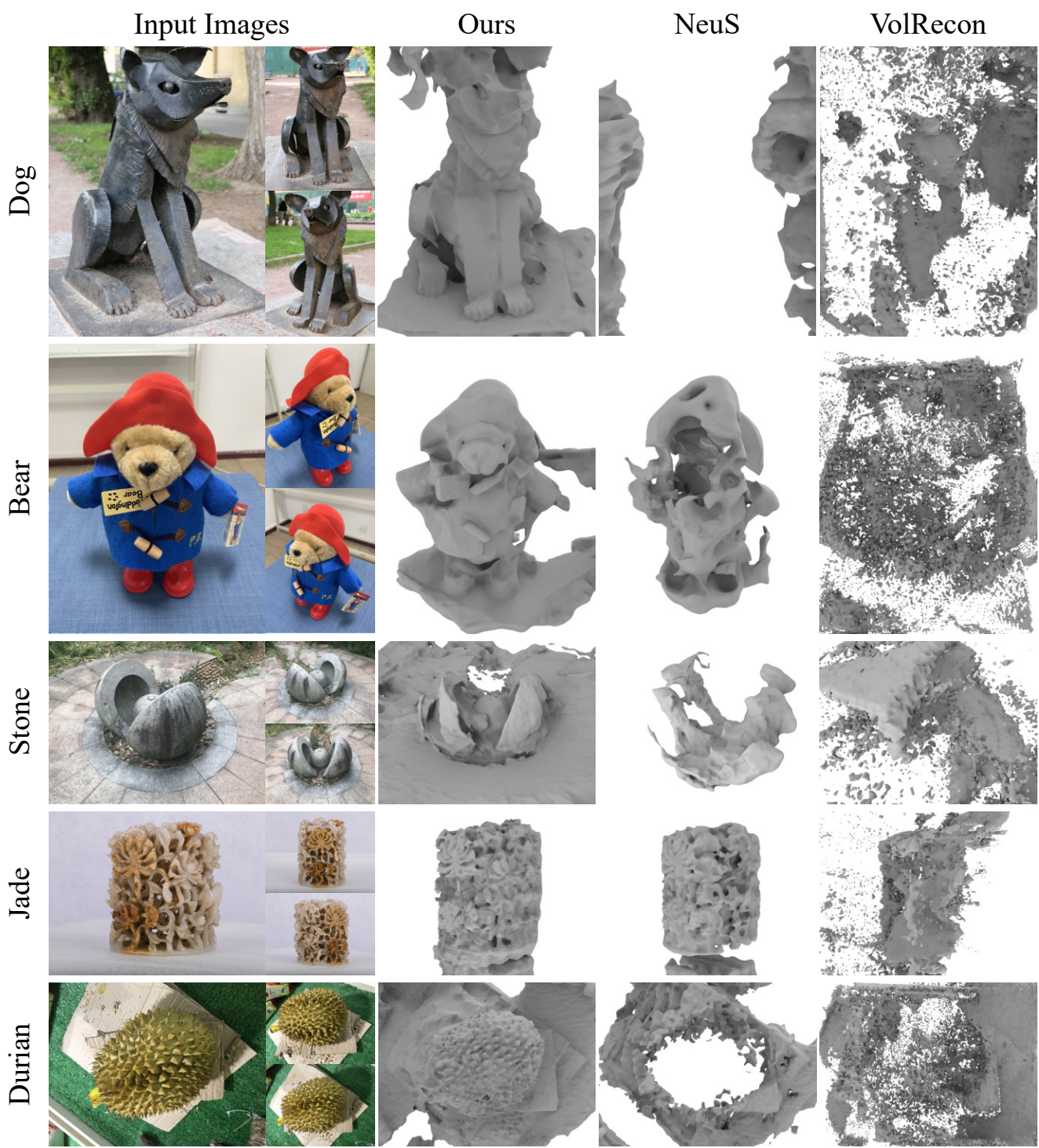

Figure 6: The remaining visual results of surface reconstruction on the random sparse setting of BlendedMVS dataset.



\end{document}